\documentclass{article} 
\usepackage{iclr2025_conference,times}

\usepackage{amsmath,amsfonts,bm}

\def\eqref#1{equation~\ref{#1}}

\def\1{\bm{1}}

\DeclareMathAlphabet{\mathsfit}{\encodingdefault}{\sfdefault}{m}{sl}
\SetMathAlphabet{\mathsfit}{bold}{\encodingdefault}{\sfdefault}{bx}{n}

\usepackage{hyperref}
\usepackage{url}
\usepackage[noabbrev, capitalise]{cleveref}
\usepackage{graphicx}
\usepackage{multirow}
\usepackage{multicol}
\usepackage{pifont}
\usepackage{wrapfig}
\usepackage{amsmath} 
\usepackage{amssymb}
\usepackage{subfig}
\usepackage{algorithm}
\usepackage{algpseudocode}
\usepackage{enumitem}

\newtheorem{theorem}{Theorem}
\newtheorem{definition}{Definition}

\usepackage{booktabs}

\definecolor{darkred}{rgb}{0.459,0.0,0.08}
\definecolor{cvprblue}{rgb}{0.21,0.49,0.74}
\hypersetup{colorlinks=true}
\hypersetup{citecolor=cvprblue}

\title{Generative Distribution Distillation}

\author{%
  Jiequan Cui \quad
  Beier Zhu \quad
  Qingshan Xu \quad
  Xiaogang Xu \quad
  Pengguang Chen \quad \\
  \textbf{Xiaojuan Qi} \quad 
  \textbf{Bei Yu} \quad
  \textbf{Hanwang Zhang} \quad 
  \textbf{Richang Hong} \quad \\
  HFUT \quad NTU \quad HKU \quad CUHK \quad SmartMore \\
  \textcolor{darkred}{\url{https://github.com/jiequancui/Generative-Distribution-Distillation}}
}

\iclrfinalcopy 
\begin{document}

\maketitle

\begin{abstract}
In this paper, we formulate the knowledge distillation (KD) as a conditional generative problem and propose the \textit{Generative Distribution Distillation (GenDD)} framework. A naive \textit{GenDD} baseline encounters two major challenges: the curse of high-dimensional optimization and the lack of semantic supervision from labels.
To address these issues, we introduce a \textit{Split Tokenization} strategy, achieving stable and effective unsupervised KD. Additionally, we develop the \textit{Distribution Contraction} technique to integrate label supervision into the reconstruction objective. 
Our theoretical proof demonstrates that \textit{GenDD} with \textit{Distribution Contraction} serves as a gradient-level surrogate for multi-task learning, realizing efficient supervised training without explicit classification loss on multi-step sampling image representations. 
To evaluate the effectiveness of our method, we conduct experiments on balanced, imbalanced, and unlabeled data. Experimental results show that \textit{GenDD} performs competitively in the unsupervised setting, significantly surpassing KL baseline by \textbf{16.29\%} on ImageNet validation set. With label supervision, our ResNet-50 achieves \textbf{82.28\%} top-1 accuracy on ImageNet in 600 epochs training, establishing a new state-of-the-art.
\end{abstract}
\section{Introduction}
\label{sec:intro}
For natural language tasks, both inputs and outputs reside in the same domain, \textit{i.e.}, language sequences, enabling the unification of diverse tasks within a single generative model optimized via next-token prediction. ChatGPT and GPT4V~\citep{ouyang2022training, achiam2023gpt} exemplify this approach with data scaling law and is often regarded as an early prototype of artificial general intelligence (AGI), showcasing the effectiveness of generative learning in natural language. Motivated by this success, researchers have begun extending generative modeling to vision and multi-modal domains~\citep{liu2023visual, li2024autoregressive, tian2024visual, zhou2024transfusion, fan2025unified, wu2024vila, yang2025mmada}, with the long-term goal of building AGI systems.

Two prominent classes of generative models have gained popularity in the vision domain: autoregressive models~\citep{li2024autoregressive, tian2024visual} and diffusion models~\citep{ho2020denoising, yang2025mmada}. Autoregressive models adopt the next-token prediction paradigm to sequentially generate image content, whereas diffusion models transform images into Gaussian noise through a forward diffusion process and learn to recover them via a reverse denoising process.
In this paper, we recast knowledge distillation (KD)~\citep{kd}, typically formulated as a discriminative task minimizing the KL divergence between categorical output distributions of the teacher and student, as a conditional generative problem modeled with the diffusion mechanism.

\begin{figure}[t!]
    \begin{minipage}{0.45\linewidth}
    \centering
    \includegraphics[width=1.0\linewidth]{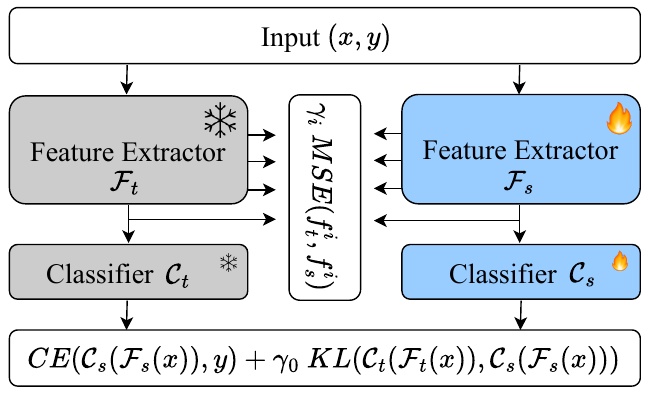}
    \caption{Previous methods are discriminative point-wise distillation.}
    \label{fig:kd}
     \end{minipage}
     \hspace{0.05in}
     \begin{minipage}{0.45\linewidth}
         \centering
         \includegraphics[width=1.0\linewidth]{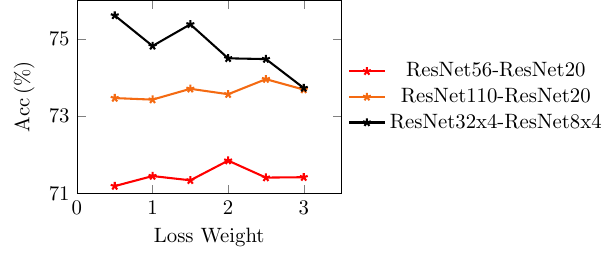}
         \caption{Sensitivity to loss weights of KD~\citep{kd}. The accuracy of student models varies with different loss weights. Optimal loss weight varies with different teacher-student configurations. }
         \label{fig:kd_loss_weight}
     \end{minipage}
     \vspace{-0.2in}
\end{figure}

Knowledge distillation (KD)~\citep{kd} has been widely adopted for knowledge transfer and model compression in real-world applications. 
As illustrated in Figure~\ref{fig:kd}, existing approaches typically guide the student model to imitate the teacher by minimizing either the KL divergence between output logits~\citep{kd, dkd, cui2024decoupled, lv2024wasserstein} or the mean squared error (MSE) between intermediate-layer features~\citep{revkd, fitnets, rkd, crd, ofd}. 
These approaches introduce additional loss terms into a multi-task framework, increasing training complexity and requiring careful loss weight tuning. Figure~\ref{fig:kd_loss_weight} presents an empirical study on CIFAR-100, showing that student performance is highly sensitive to the choice of the loss weight. Moreover, the optimal weight also differs across teacher-student configurations, underscoring the limited robustness and generalizability of these methods in diverse application scenarios. This problem could exacerbate as the number of loss weight increases.

{\bf Generative Distribution Distillation (GenDD).}
Inspired by the success of generative learning~\citep{li2024autoregressive, ho2020denoising, song2020score}, we leverage diffusion models to 
formulate KD as a single generative process. 
\begin{wrapfigure}{r}{0.44\textwidth}
    \centering
    \includegraphics[width=.85\linewidth]{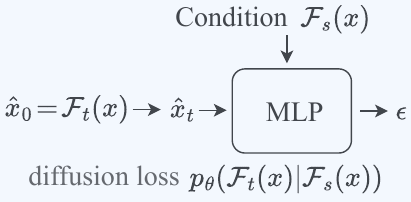}
    \caption{Conditional generation for KD.}
    \label{fig:diffloss}
\end{wrapfigure}
As shown in Figure~\ref{fig:diffloss}, taking image representation $\mathcal{F}_{s}(x)$ of the student model as conditional inputs, we learn to generate the representation $\mathcal{F}_{t}(x)$ of the teacher model, thus achieving distribution mapping between the student and teacher. \quad\quad\quad\quad\quad\quad
{\bf Challenges of GenDD.} MAR~\citep{li2024autoregressive} deploys a diffusion loss for autoregressive image generation. Specifically, images are tokenized in continuous token sequences via VAE~\citep{kingma2013auto} and then fed into autoregressive models. However, image tokens from VAE 
only have a dimension of 16 while image representation in classification often 
have a large dimension, reaching to 2048. We empirically observe \textit{the high-dimensional optimization disaster: the training can't converge or even crashes}. Moreover, diffusion models are optimized by variational lower bound (VLB) to reconstruct inputs, which \textit{lacks semantic constraints with labels and thus hinders model performance.}.

{\bf Our Solution.} To tackle the high-dimensional optimization challenge, we propose the \textit{Split Tokenization}: split image representation $\mathcal{F}_{t}(x)$ into token sequences with positional index.
Conditoned on $\mathcal{F}_{s}(x)$, models are trained to reconstruct these tokens individually. Such a \textit{Split Tokenization} strategy effectively stabilize the training of \textit{GenDD}, achieving unsupervised KD. To enable label supervision of \textit{GenDD}, we develop a \textit{Distribution Contraction} technique. We theoretically prove that \textit{GenDD} with \textit{Distribution Contraction} serves as a surrogate to multi-task learning (combining reconstruction and classification loss), eliminating explicit classification loss and multi-step sampling and thus leading to efficient and effective supervised KD. 

{\bf Our Results.}
To validate the effectiveness of our method, we conduct experiments on balanced data including CIFAR-100~\citep{cifar} and ImageNet~\citep{imagenet}, imbalanced data like long-tailed ImageNet~\citep{liu2019large}, and unlabeled data CC3M~\citep{changpinyo2021conceptual}.
With the \textit{Split Tokenization}, our \textit{GenDD} model significantly surpasses the KL baseline by \textbf{16.29\%} in the unsupervised KD setting. Moreover, with \textit{Distribution Contraction technique}, \textit{GenDD} incorporating label supervision largely outperforms previous distillation methods. Especially, we achieve new state-of-the-art ResNet-50 performance on ImageNet.

In summary, our contributions are as follows:
\begin{itemize}[leftmargin=15pt]
    \item We formulate KD as a conditional generation problem and propose the \textit{Generative Distribution Distillation (GenDD)} algorithm.
    \item To address the high-dimensional optimization challenge, we propose a \textit{Split Tokenization} strategy. To enable label supervision of \textit{GenDD}, the \textit{Distribution Contraction} technique is developed.
    \item We theoretically prove that \textit{GenDD} with \textit{Distribution Contraction} acts as a gradient-level surrogate for the multi-task learning, resulting in effecient and effective optimization.
    \item We empirically show the advantage of our method on balanced, imbalanced, and unlabeled data. Specifically, we achieve the state-of-the-art performance on ImageNet for both unsupervised and supervised KD settings.
\end{itemize}

\begin{table}[h]
    \centering
    \caption{Comparison with previous distillation methods.}
    \resizebox{1.0\linewidth}{!}{
    \begin{tabular}{cccc}
       \toprule
       Method& Generative or Discriminative &Distributional or Point-wise &Sensitivity to Loss Weight \\
       \midrule
        Logits-based  &discriminative &point-wise &sensitive \\
        Feature-based &discriminative &point-wise &sensitive \\
        \midrule
         \textbf{GenDD} &generative &distributional &NA\\
        \bottomrule
    \end{tabular}
    }
    \label{tab:comparison_related_work}
\end{table}

\section{Related Work}
\label{sec:related_work}

{\bf Generative Learning.}
In NLP, generative modeling based on next-token prediction (NTP) forms the foundation of the GPT series of language models~\citep{radford2018improving, radford2019language, brown2020language, ouyang2022training, achiam2023gpt}. This approach enables unsupervised learning from large-scale corpora and has driven significant advances in zero-shot and few-shot generalization, culminating in powerful systems such as ChatGPT and GPT4V~\citep{ouyang2022training, achiam2023gpt} that exhibit strong performance across diverse natural language tasks without task-specific fine-tuning. Subsequently, autoregressive (AR) models leveraging NTP have also gained popularity in vision~\citep{li2024autoregressive, tian2024visual} and multi-modal~\citep{liu2023visual, zhu2023minigpt} domains, fostering the development of unified generalist models capable of both understanding and generation~\citep{zhou2024transfusion, fan2025unified, wu2024vila, yang2025mmada} on multi-modal data.

Besides NTP and autoregressive (AR) models, diffusion models~\citep{sohl2015deep, song2019generative, ho2020denoising, song2020score} have emerged as a powerful class of generative methods, demonstrating impressive sample quality and robustness~\citep{rombach2022high, yang2025mmada, zhou2024transfusion}. However, they often require multi-step iterative sampling during inference, which can be computationally expensive and time-consuming. Recently, flow matching approaches~\citep{geng2025mean,lipman2022flow, gat2024discrete} have been proposed to address these limitations by providing efficient and scalable generative modeling with fewer sampling steps while maintaining high fidelity.

{\bf Knowledge Distillation.}
Knowledge distillation (KD)~\citep{kd} is developed to transfer ``dark knowledge'' from a teacher model to a student model. The core idea is to leverage the soft targets—\textit{i.e.}, the probability distribution over classes—produced by the teacher to guide the training of the student. These soft labels contain rich information about inter-class similarities that are not captured by one-hot labels, thereby enabling the student to learn more generalizable representations. Since the success of KD~\citep{kd}, advanced logit-based~\citep{furlanello2018born, zhang2018deep, cho2019efficacy, huang2022knowledge, dkd, hao2023vanillakd, cui2024decoupled, cui2025generalized} and feature-based~\citep{fitnets,rkd, crd, ofd, revkd, huang2023knowledge} algorithms have been proposed. However, these complicated methods are often sensitive to loss weights and require hyperparameter tuning for each teacher-student configuration.

Additionally, previous KD methods are typically trained along with discriminative cross-entropy loss and promote consistency between the teacher and student on each data point. In this paper, the proposed \textit{GenDD} is optimized with a single reconstruction objective and models the distribution of each example. Refer to Table~\ref{tab:comparison_related_work} for the comparison summary with previous work.

\section{Method}
\label{sec:method}

\subsection{Revisiting Knowledge Distillation as Multi-task Learning}
A typical classification model comprises a feature extractor $\mathcal{F}(\cdot)$ and a classifier $\mathcal{C}(\cdot)$. Given an input image $x$, the model produces a feature representation $\mathcal{F}(x)$ and a corresponding prediction $\arg\max \mathcal{C}(\mathcal{F}(x))$. While knowledge distillation (KD) is broadly applicable to a wide range of real-world scenarios, we focus on image classification in this work.
KD is designed to transfer the inductive knowledge of the teacher model to the student model, enabling both model compression and improved generalization. Previous KD methods~\citep{kd,crd,revkd} are often discriminative and point-wise, minimizing KL divergence or Mean Square Error (MSE) between sample logits or intermediate-layer features,
\begin{equation}
    \min_{\theta} CE(\mathcal{C}_{s} \circ \mathcal{F}_{s}(x), y) + \gamma_{0} \cdot KL(\mathcal{C}_{t} \circ \mathcal{F}_{t}(x), \mathcal{C}_{s} \circ \mathcal{F}_{s}(x)) + \sum_{i=1} \gamma_{i} \cdot MSE(f_{t}^{i}, f_{s}^{i}), 
    \label{eq:kd_multi_task}
\end{equation}
where $\theta$ is parameters of student, $\{\gamma_{i}\}$ are hyper-parameters for multi-task learning.

As illustrated in Figure~\ref{fig:kd_loss_weight}, the performance of the student model is notably sensitive to the choice of hyperparameters, even on the \textit{same training dataset} with different teacher-student configurations, making the optimization challenging. For more detailed comparisons with advanced KD methods, please refer to Appendix~\ref{sec:comparison_hyper_parameter}. 
On the other hand, in real-world scenarios, the teacher can be trained on custom data that can't be accessed publicly because of \textit{privacy protection}. In this case, previous algorithms can't work well without cross-entropy in~\cref{eq:kd_multi_task}, which is validated in Section~\ref{sec:gendd_without_label}.

In contrast, our proposed \textbf{Generative Distribution Distillation (GenDD)} is optimized by a unified \textit{reconstruction objective}, eliminating the need for extensive hyperparameter tuning. Furthermore, it achieves competitive performance using only unlabeled data (Section~\ref{sec:gendd_without_label}) and attains state-of-the-art results when annotation supervision is available (Section~\ref{sec:gendd_with_label}).

\subsection{Generative Distribution Distillation (GenDD)}
\label{sec:gendd}

Inspired by the success of ChatGPT~\citep{ouyang2022training, achiam2023gpt} in natural language processing (NLP), recent efforts have aimed to unify multi-modal understanding and generation within a single generative framework~\citep{zhou2024transfusion, wu2024vila, yang2025mmada, fan2025unified}. In particular, diffusion models have recently emerged as promising alternatives to large language models (LLMs)~\citep{arriola2025block, yang2025mmada}.
In this paper, we propose to formulate knowledge distillation (KD) as a generative learning process based on diffusion foundations.

\subsubsection{GenDD with Unlabeled Data}
\label{sec:gendd_without_label}

Given an image $x \in X$, $\hat{x}_{0} \sim q(\mathcal{F}_{t}(x))$, taking $\mathcal{F}_{s}(x)$ as condition, we learn to reconstruct the image representation of teacher model, \textit{i.e.}, $\hat{x}_{0}$, with the following training objective,
\begin{equation}
    \mathbb{E}_{x, m, \epsilon} \left[ ||\epsilon - \epsilon_{\theta}(\hat{x}_{m}, m, \mathcal{F}_{s}(x)) ||^{2} \right],
    \label{eq:gendd}
\end{equation}
where $\epsilon \in \mathcal{N}(\mathbf{0}, \mathbf{I})$, $m \in [0, M]$ is the sampled time step ($M$ is the maximum), $\hat{x}_{m} = \sqrt{\bar \alpha_{m}} \hat{x}_{0} + \sqrt{1-\bar \alpha_{m}} \epsilon$ is the noisy input at time step $m$, in particular, $\hat{x}_{0} = \mathcal{F}_{t}(x)$, $\bar \alpha_{m}=\Pi_{i=1}^{m} \alpha_{i}$, and $\alpha$ is defined with a variance schedule~\citep{ho2020denoising, nichol2021improved}.

At inference, with an input image $x$ and a sampled $\hat{x}_{M}^{'} \in \mathcal{N}(\mathbf{0}, \mathbf{I})$, the image representation $\hat{x}_{0}^{'}$ could be generated through iterative update from $m=M$ to $m=0$:
\begin{equation}
    \hat{x}_{m-1}^{'} = \frac{1}{\sqrt{\alpha_{m}}} \left( \hat{x}_{m}^{'} - \frac{1-\alpha_{m}}{\sqrt{1-\bar \alpha_{m}}} \epsilon_{\theta}(\hat{x}_{m}^{'}, m, \mathcal{F}_{s}(x))\right) + \sigma_{m} \epsilon,
\end{equation}
where $\epsilon \in \mathcal{N}(\mathbf{0}, \mathbf{I})$, $\sigma_{m}$ could be learned or pre-defined. Note that, the reverse diffusion process could be respaced~\citep{li2024autoregressive, song2020denoising} for efficient sampling.

Then, the final prediction could be derived by inputting the image representation $\hat{x}_{0}^{'}$ into teacher model classifier, \textit{i.e.}, $\mathop{\arg\max} \mathcal{C}_{t}(\hat{x}_{0}^{'})$.

{\bf High-dimensional Optimization Disaster.}
Following MAR~\citep{li2024autoregressive}, we implement the diffusion head using a 3-layer MLP. MAR~\citep{li2024autoregressive} showcases the effectiveness of continuous tokenizers for autoregressive image generation, where images are first tokenized into sequences of continuous tokens using a VAE~\citep{kingma2013auto}. These tokens are then fed into an autoregressive model that learns the per-token distribution through a diffusion loss. 

However, the dimensionality of each token in VAE~\citep{kingma2013auto} is limited to 16, whereas feature representations in image classification tasks typically have much higher dimensionality, reaching up to 2048. 
Our empirical study reveals a high-dimensional optimization issue, particularly when the feature dimension of the student model, $\mathrm{Dim}(\mathcal{F}_s(x))$, is much lower than that of the teacher model, $\mathrm{Dim}(\mathcal{F}_t(x))$, often leading to training instability or failure. Refer to Section~\ref{sec:exp_ablation} for more details.

{\bf Split Tokenization.}
To address the challenges of high-dimensional optimization, we propose decomposing the feature representation $\hat{x}_{0}$ into a sequence of lower-dimensional tokens. Specifically, we define the \textit{SplitTok} operation as:
\begin{equation}
\textit{SplitTok}(\mathcal{F}_{t}(x)) = \left[(\hat{x}_{0}^{1}, 1, \mathcal{F}_{s}(x)), (\hat{x}_{0}^{2}, 2, \mathcal{F}_{s}(x)), \dots, (\hat{x}_{0}^{n}, n, \mathcal{F}_{s}(x))\right],
\end{equation}
where each tuple consists of a token $\hat{x}_{0}^{i}$, its position index $i$, and the conditioning context from the student model $\mathcal{F}_{s}(x)$. Based on this structure, we reformulate the training objective in Eq.~\ref{eq:gendd} into a token-wise form:
\begin{equation}
\mathbb{E}_{y, \text{id}, c, m, \epsilon} \left[ || \epsilon - \epsilon_{\theta}(\hat{y}_{m}, m, \text{id}, c) ||^{2} \right],\quad where\ (y, \text{id}, c) \sim q\left(\textit{SplitTok}(\mathcal{F}_{t}(x))[\text{id}]\right).
\label{eq:gendd_split}
\end{equation}
This token-based formulation allows the model to operate in lower-dimensional subspaces, thereby mitigating instability during optimization in high-dimensional feature spaces.

\subsubsection{\bf GenDD with Label Supervision}
\label{sec:gendd_with_label}
Conditioned on the student’s feature representation, \textit{GenDD} reconstructs the teacher’s feature tokens, enabling unsupervised knowledge distillation. However, the reconstruction objective alone fails to exploit label supervision during training. To address this problem, we introduce a \textit{Distribution Contraction} mechanism that enables \textit{GenDD} to effectively incorporate label information into the optimization process.

{\bf Multi-task Learning.}
To incorporate label supervision, a straightforward baseline is multi-task learning, which combines the reconstruction objective with a standard cross-entropy loss:
\begin{equation}
\begin{aligned}
\min_{\theta_s} ; & \mathcal{L}_{\text{CE}} = - y \log \mathcal{C}_{s}(\hat{x}_{0}^{'}), \\
\text{s.t.} & \min_{\theta}  \mathbb{E}_{y, \text{id}, c, m, \epsilon} \left[ \left| \epsilon - \epsilon_{\theta}(\hat{y}_m, m, \text{id}, c) \right|^2 \right],
\label{eq:gendd_multi_task}
\end{aligned}
\end{equation}
where $\mathcal{C}_s$ denotes the classifier on top of the reconstructed representation $\hat{x}_{0}^{'}$, $\theta=(\theta_{s},\theta_{diff})$ are parameters of student model and diffusion head respectively.

As shown in~\cref{eq:gendd_multi_task}, the cross-entropy loss encourages the generated feature $\hat{x}_{0}^{'}$ to be correctly classified, while the reconstruction loss regularizes the student feature space to align with that of the teacher. These two objectives can be optimized either alternately or simultaneously during training. 

However, in practice, involving $\hat{x}_{0}^{'}$ directly in the training of the diffusion model is computationally inefficient. Since $\hat{x}_{0}^{'}$ must be sampled through a multi-step reverse diffusion process at each iteration, using it as an intermediate target for supervision substantially increases training time and resource consumption. Moreover, gradients cannot be efficiently propagated through the sampling chain, limiting the effectiveness of end-to-end optimization.

{\bf GenDD with Label Supervision.}
Instead of relying on conventional multi-task learning, we incorporate label supervision through the proposed \textit{Distribution Contraction} technique, formally defined in Definition~\ref{thm:distribution_contraction}. Furthermore, Theorem~\ref{thm:thm_gendd} establishes that \textit{GenDD}, trained with the \textit{Distribution Contraction} technique, serves as an efficient and effective surrogate for multi-task learning.

\begin{definition}[Distribution Contraction]
\label{thm:distribution_contraction}
\normalfont
Let $\hat{x}_{0} \sim q(\mathcal{F}_t(x))$ be the feature representation produced by a well-trained teacher model $\mathcal{F}_t$ on input $x \in X$, where $\mathcal{C}_t(\hat{x}_0) \in \mathbb{R}^C$ is the softmax-based class probability vector over $C$ categories. To incorporate label supervision $y$ during diffusion model training, we enhance the semantic consistency of $\hat{x}_{0}$ by contracting it toward the class center $c_y$:
\begin{equation}
    \tilde{x}_0 = \lambda \hat{x}_0 + (1 - \lambda) c_y,
    \label{eq;distribution_contraction}
\end{equation}
where $\lambda \in [0, 1]$ controls the degree of contraction, $c_y$ denotes the centroid of features for class $y$.
\end{definition}

\begin{theorem}[Distribution Contraction Approximates Multi-task Learning at Gradient Level]
\label{thm:thm_gendd}
Assume the teacher model, composed of a feature extractor $\mathcal{F}_{t}(\cdot)$ and a linear classifier $\mathcal{C}_{t}(\cdot)$, is well-trained, then the optimization of the reconstruction objective with distribution contraction in Definition~\ref{thm:distribution_contraction}:
\begin{equation}
    \mathcal{L}_{GenDD} = \mathbb{E}_{x, m, \epsilon} \left[ \left\| \epsilon - \epsilon_{\theta}(\tilde{x}_{m}, m, \mathcal{F}_{s}(x)) \right\|^{2} \right], 
    \label{eq:gendd_label}
\end{equation}
is approximately equivalent, at the gradient level, to optimizing the multi-task objective:
\begin{equation}
    \mathcal{L}_{multi} = \gamma_{0} \ \mathbb{E}_{x, m, \epsilon} \left[ \left\| \epsilon - \epsilon_{\theta}(\hat{x}_{m}, m, \mathcal{F}_{s}(x)) \right\|^{2} \right] 
    + 
    \gamma_{1} \ \mathbb{E}_{x} \left[ \mathcal{L}_{\mathrm{CE}} \left( \mathcal{C}_{t}(\hat{x}_{0}^{\prime}), y \right) \right],
    \label{eq:multi_task_parallel}
\end{equation}
where $\gamma_{0}$ and $\gamma_{1}$ are constants controlling the relative weights of the two loss terms, $\mathcal{C}_{t}(\cdot)$ is frozen. 

\textit{Proof.} See Appendix~\ref{sec:proof_theorem1}.
\end{theorem}

{\bf Remark.} Theorem~\ref{thm:thm_gendd} shows that $\mathcal{L}_{\text{GenDD}}$ acts as a gradient-level surrogate for the multi-task objective $\mathcal{L}_{\text{multi}}$, avoiding explicit classification loss optimization and eliminating the need for multi-step sampling to obtain $\hat{x}_{0}^{'}$ during training. This enables more efficient training while retaining strong performance with label supervision.

\section{Experiments}
\label{sec:exp}
Section~\ref{sec:exp_gendd_unsup} presents the competitive performance of \textit{GenDD} under the unsupervised KD setting. When label supervision is incorporated via \textit{Distribution Contraction}, \textit{GenDD} achieves strong results on both balanced and imbalanced datasets, as shown in Section~\ref{sec:exp_gendd_sup}. Finally, we perform ablation studies in Section~\ref{sec:exp_ablation} to assess the impact of the proposed \textit{Split Tokenization} and \textit{Distribution Contraction} techniques.

{\bf Experimental Settings.}
Following prior work~\citep{li2024autoregressive}, we implement the diffusion head using a 3-layer MLP. During training, the maximum diffusion step is set to $M=1000$. At inference time, we apply a 64-step sampling procedure to generate the feature representation $\hat{x}_{0}^{'}$. For \textit{Split Tokenization}, the feature representation $\hat{x}_{0}$ is divided into non-overlapping tokens, each with a dimensionality of 64. To enhance generation quality, we employ classifier-free guidance with a scale of $2.0$.

For the unsupervised KD setting, we evaluate \textit{GenDD} on the target dataset, \textit{i.e.}, ImageNet~\citep{deng2009imagenet}, and non-target dataset CC3M~\citep{changpinyo2021conceptual}. Under the supervised KD setting, we train various teacher-student configurations on both balanced (including ImageNet~\citep{deng2009imagenet} and CIFAR~\citep{krizhevsky2009learning}) and imbalanced data (ImageNet-LT~\citep{liu2019large}).

\subsection{GenDD in Unsupervised Setting}
\label{sec:exp_gendd_unsup}
To evaluate the effectiveness of \textit{GenDD} for unsupervised knowledge distillation (KD), we train models on both target data (ImageNet) and non-target data (CC3M), and assess their performance on the ImageNet validation set. The results are summarized in Table~\ref{tab:gendd_unlabeled}.

For teacher-student configurations such as (ResNet-34, ResNet-18) and (ResNet-50, MobileNet), we adopt pre-trained teacher models from PyTorch. Since these teachers have been trained on the target dataset, their predictions closely approximate the ground-truth, allowing conventional KL-based distillation without cross-entropy to perform competitively relative to \textit{GenDD}.

However, in practical unsupervised KD scenarios, custom training data, along with their annotations, can be both private and inaccessible. To simulate this setting, we train student models on non-target data, specifically CC3M~\citep{changpinyo2021conceptual}, where teacher models have never been exposed to the data. In this case, teacher predictions become less reliable, and naive KL-based distillation without cross-entropy for label supervision fails to produce satisfactory results. As shown in Table~\ref{tab:gendd_unlabeled}, with \textit{GenDD}, our MobileNet achieves 67.89 top-1 accuracy, significantly outperforms the KL baseline by \textbf{16.29\%}. 

\begin{table*}[tb!]
\center
\caption{\textbf{Top-1 accuracy~(\%) on the ImageNet validation with supervised GenDD}. All results are the average over three trials. ``*" represents that the models are reproduced with the cosine learning rate schedule for fair comparison.}
\resizebox{1.0\linewidth}{!}
{
\begin{tabular}{ccccccccccc}
\toprule
 &\multirow{2}{*}{Teacher} &\multirow{2}{*}{Student}  &\multicolumn{7}{c}{Discriminative Point-wise Distillation} &Gen.D.D.  \\
 \cmidrule{4-11}
 & & &AT &OFD &CRD &ReviewKD &DKD* &IKL-KD* &KD*  &\textbf{GenDD} \\ 
 \midrule
 \multicolumn{11}{c}{\textit{ResNet-34}, \textit{ResNet-18}, \textit{Regular recipe}, \textit{100 epochs}} \\
 \midrule
 Top-1 &73.31 &69.75 &70.69 &70.81 &71.17 &71.61 &71.87 &71.91 &71.24 &\textbf{72.38} \\
 Top-5 &91.42 &89.07 &90.01 &89.98 &90.13 &90.51 &90.45 &90.52 &90.23  &\textbf{90.63}\\
 \midrule
 \multicolumn{11}{c}{\textit{ResNet-50}, \textit{MobileNet}, \textit{Regular recipe}, \textit{100 epochs}} \\
 \midrule
 Top-1 &76.16 &68.87 &69.56 &71.25 &71.37 &72.56 &72.55 &73.19 &71.44  &\textbf{73.78} \\
 Top-5 &92.86 &88.76 &89.33 &90.34 &90.41 &91.00 &91.05 &91.47 &90.35  &\textbf{91.56} \\
\midrule
\multicolumn{11}{c}{\textit{BEiTv2}, \textit{ResNet-50}, \textit{Strong recipe}, \textit{300 (A2) or 600 (A1) epochs}} \\
\midrule
Top-1 (BEiT-L-A2) &88.01 &79.80 &- &- &79.48 &79.11 &80.77 &80.98 &80.89  &\textbf{81.64} \\
Top-1 (BEiT-B-A2) &86.12 &79.80 &- &- &- &- &- &- &80.96  &\textbf{81.76} \\
Top-1 (BEiT-L-A1) &88.01 &80.38 &- &- &- &- &81.83 &- &81.68  &\textbf{82.28} \\
\bottomrule
\end{tabular}
}   
\label{tab:imagenet_gendd_label}
\end{table*}

\begin{table}[tb!]
  \centering
  \begin{minipage}{0.48\linewidth}
    \centering
    \caption{\textbf{Top-1 accuracy(\%) on the ImageNet validation with unsupervised GenDD.}}
    \resizebox{1.0\linewidth}{!}
    {
    \begin{tabular}{lccc}
      \toprule
      Method &Teacher &Student &Accuracy \\
      \midrule
      \multicolumn{4}{c}{w/o Label On Target Data, \textit{i.e.}, ImageNet-1K} \\
      \midrule
      KL    &ResNet-50 &MobileNet &71.40 \\
      \textbf{GenDD} &ResNet-50 &MobileNet &\textbf{72.03} \\
      \midrule
      \multicolumn{4}{c}{w/o Label On Non-target Data, \textit{i.e.}, CC3M} \\
      \midrule
      KL &ResNet-50 &MobileNet &51.60 \\
      \textbf{GenDD} &ResNet-34 &ResNet-18 &\textbf{66.90} \\
      \textbf{GenDD} &ResNet-50 &MobileNet &\textbf{67.89} \\
      \bottomrule
    \end{tabular}
    }
    \label{tab:gendd_unlabeled}
  \end{minipage}
  \hfill
  \begin{minipage}{0.45\linewidth}
    \centering
    \caption{\textbf{Top-1 accuracy(\%) on the ImageNet-LT validation with GenDD.} ``*'' represents the unsupervised setting.}
    \resizebox{1.0\linewidth}{!}
    {
    \begin{tabular}{lccc}
      \toprule
      Method & Teacher &Student &Accuracy \\
      \midrule
      Baseline &- &ResNet-18 &41.15 \\
      Baseline &- &ResNet-50 &45.47 \\
      \midrule
      KD &ResNeXt-101 &ResNet-18 &44.32 \\
      KD &ResNeXt-101 &ResNet-50 &48.31 \\
      IKL-KD &ResNeXt-101 &ResNet-18 &45.21 \\
      IKL-KD &ResNeXt-101 &ResNet-50 &49.29 \\
      \midrule
      \textbf{GenDD*} &ResNext-101 &ResNet-18 &\textbf{45.54} \\
      \textbf{GenDD*} &ResNeXt-101 &ResNet-50 &\textbf{49.31} \\
      \bottomrule
    \end{tabular}
    }
    \label{tab:imagenetlt}
  \end{minipage}
\end{table}

\begin{table*}[tb!]
\center
\caption{\textbf{Top-1 accuracy~(\%) on the CIFAR-100 validation.} Teachers and students are in the \textbf{same} architectures. $\Delta$ represents the improvements over the KD~\citep{kd} baseline. All results are the average over three trials.}
\resizebox{1.0\linewidth}{!}
{
\begin{tabular}{cccccccc}
\toprule
\multirow{4}{*}{\begin{tabular}[c]{@{}c@{}}Distillation \\ Manner\end{tabular}} & \multirow{2}{*}{Teacher}  & ResNet56 & ResNet110 & ResNet32$\times$4 & WRN-40-2 & WRN-40-2 & VGG13 \\
&      & 72.34      & 74.31       & 79.42        & 75.61      & 75.61      & 74.64   \\
& \multirow{2}{*}{Student}  & ResNet20 & ResNet32 & ResNet8$\times$4  & WRN-16-2 & WRN-40-1 & VGG8  \\
& \space     & 69.06      & 71.14       & 72.50        & 73.26      & 71.98      & 70.36   \\
\midrule
\multirow{8}{*}{\shortstack{Discriminative \\ Point-wise \\ Distillation}}
& FitNet   &69.21   &71.06   &73.50   &73.58    &72.24    &71.02   \\
& RKD      &69.61   &71.82   &71.90   &73.35    &72.22    &71.48   \\
& CRD      &71.16   &73.48   &75.51   &75.48    &74.14    &73.94   \\
& OFD      &70.98   &73.23   &74.95   &75.24    &74.33    &73.95   \\
& ReviewKD &71.89   &73.89   &75.63   &76.12    &75.09    &74.84   \\ 
\cmidrule{2-8}                                                       
& DKD      &71.97   &74.11   &76.32   &76.24    &74.81    &74.68 \\
& IKL-KD   &71.44   &74.26   &76.59   &76.45    &74.98    &\textbf{74.98} \\
& KD       &70.66   &73.08   &73.33   &74.92    &73.54    & 72.98   \\
\midrule
\multirow{2}{*}{Gen.D.D.} 

&\textbf{GenDD}                &\textbf{72.63} &\textbf{74.95} &\textbf{77.47} &\textbf{76.83} &\textbf{75.98} &74.24 \\
& $\Delta$             &\textbf{+1.97}  &\textbf{+1.87}  &\textbf{+4.14} &\textbf{+1.91} &\textbf{+2.44} &\textbf{+1.26} \\
\bottomrule
\end{tabular}
}
\label{tab:cifar_kd_1}
\end{table*}

\begin{table*}[tb!]
\center
\caption{\textbf{Top-1 accuracy~(\%) on the CIFAR-100 validation.} Teachers and students are in \textbf{different} architectures. All results are the average over 3 trials.}
\resizebox{1.0\linewidth}{!}
{
\begin{tabular}{ccccccc}
\toprule
\multirow{4}{*}{\begin{tabular}[c]{@{}c@{}}Distillation \\ Manner\end{tabular}} & \multirow{2}{*}{Teacher}  & ResNet32$\times$4 & WRN-40-2 & VGG13 & WRN-40-2 & ResNet32$\times$4 \\
& \space     & 79.42      & 75.61       & 74.64        & 75.61      & 79.42         \\
& \multirow{2}{*}{Student}  & ShuffleNet-V1 & ShuffleNet-V1 & ResNet20  &ResNet8x4 & ShuffleNet-V2  \\
& \space     &70.50      &70.50       &69.06        &72.50      &71.82         \\ 
\midrule
\multirow{8}{*}{\shortstack{Discriminative \\ Point-wise \\ Distillation}}
& FitNet     &73.59   &73.73   &69.27   &74.74   &73.54    \\
& RKD        &72.28   &72.21   &69.83   &72.43   &73.21    \\
& CRD        &75.11   &76.05   &71.16   &76.64   &75.65    \\
& OFD        &75.98   &75.85   &-   &74.50   &76.82    \\
& ReviewKD   &77.45   &77.14   &-   &75.48   &77.78    \\     
\cmidrule{2-7}                                                       
& DKD        &76.45   &76.70   &70.57   &75.23   &77.07       \\
& IKL-KD     &76.64   &\textbf{77.19}   &70.88   &76.12   &77.16       \\
& KD         &74.07   &74.83   &70.20   &74.01   &74.45       \\
\midrule
\multirow{2}{*}{Gen.D.D.}
&\textbf{GenDD}               &\textbf{77.58} &76.73 &\textbf{72.17} &\textbf{77.55} &\textbf{78.13} \\
& $\Delta$           & \textbf{+3.51} &\textbf{+1.90} &\textbf{+1.97} &\textbf{+3.54} &\textbf{+3.68} \\
\bottomrule
\end{tabular}
}
\label{tab:cifar_kd_2}
\end{table*}

\subsection{GenDD with Label Supervision}
\label{sec:exp_gendd_sup}
Our empirical study on balanced data, including CIFAR and ImageNet, is presented in Section~\ref{sec:exp_balanced_data}. Section~\ref{sec:exp_imbalanced_data} discusses the effects of \textit{GenDD} on imbalanced data, \textit{i.e.}, ImageNet-LT.

\subsubsection{Experimental Results on Balanced Data}
\label{sec:exp_balanced_data}

{\bf Experimental Results on ImageNet.}
On ImageNet, we evaluate a range of teacher-student configurations, covering diverse network architectures (CNNs with regular or depth-wise convolutions, and Transformers), training recipes (standard vs. strong augmentation), and model scales (e.g., ResNet-34, ResNet-50, BEiT-Large).
\textit{Despite the significant variation across configurations, we employ a consistent $\lambda=0.9$ for the \textit{Distribution Contraction} in Definition~\ref{thm:distribution_contraction}, highlighting the generalizability, robustness, and practical convenience of GenDD.}

Under the regular training recipe (including \textit{RandomResizedCrop} and horizontal flip), we train models 100 epochs with a cosine learning rate schedule. For fair comparisons, we reproduce the results of KD, DKD, and IKL-KD with their open-sourced code and just replace the step learning rate schedule with the cosine learning rate schedule. Our ResNet-18 achieves a top-1 accuracy of 72.38\%, outperforming KD, IKL-KD, and DKD by 1.14\%, 0.47\%, and 0.51\%, respectively. Similarly, our MobileNet reaches 73.78\% top-1 accuracy, surpassing KD, IKL-KD, and DKD by 2.34\%, 0.59\%, and 1.23\%, respectively.

When applying a strong training recipe, prior work~\citep{hao2023vanillakd} shows that recent advanced KD methods such as DKD and ReviewKD only perform comparably to the original KD:
\begin{itemize}
    \item A2: \textit{RandAug(7/0.5)}, \textit{MixUp}: 0.1, \textit{CutMix}: 1.0, \textit{Label Smoothing}: 0.0, training 300 epochs.
    \item A1: \textit{RandAug(7/0.5)}, \textit{MixUp}: 0.2, \textit{CutMix}: 1.0, \textit{Label Smoothing}: 0.1, training 600 epochs.
\end{itemize}

Remarkably, our \textit{GenDD} models consistently outperform these baselines by a significant margin with the same training settings. Specifically, taking BEiTv2-Large as the teacher, our ResNet-50 achieves 82.28\% top-1 accuracy with the A1 training recipe.

{\bf Experimental Results on CIFAR.}
Following previous work~\citep{cui2024decoupled,revkd}, we consider the distillation among the architectures having the same unit structures, like ResNet56 and ResNet20, VGGNet13 and VGGNet8. On the other hand, we also explore the distillation among architectures made up of different unit structures, like WideResNet and ShuffleNet, VGGNet and ResNet. Specifically, we train all models for 240 epochs with a learning rate that decays by 0.1 at the 150th, 180th, and 210th epoch.

Experimental results on CIFAR-100 are summarized in Table~\ref{tab:cifar_kd_1} and Table~\ref{tab:cifar_kd_2}. Table~\ref{tab:cifar_kd_1} lists the comparisons with previous methods
under the setting that the architectures of the teacher and student have the same unit structures.
As shown in Table~\ref{tab:cifar_kd_1}, \textit{GenDD} models can achieve much better or comparable performance in all considered settings. Specifically, we achieve the best performance in 5 out of 6 training settings. With teacher-student configurations of (ResNet-110, ResNet32) and (ResNet-56, ResNet20), we achieve 72.63\% and 74.95\% top-1 accuracy respectively. 

Table~\ref{tab:cifar_kd_2} lists the comparisons with previous methods under the setting that the architectures of the teacher and student have different unit structures. As shown in Table~\ref{tab:cifar_kd_2}, we achieve the best performance in 4 out of 5 training configurations.

\subsubsection{Experimental Results on Imbalanced Data}
\label{sec:exp_imbalanced_data}
Real-world data often exhibits a long-tailed distribution, making long-tailed recognition a critical challenge for practical applications. Extensive research has been devoted to addressing this problem through both algorithmic and theoretical advances~\citep{cui2019class,cao2019learning, kang2019decoupling, cui2022reslt, menon2020long, cui2021parametric, cui2023generalized, cui2024classes}. Following recent efforts~\citep{cui2024decoupled, cui2025generalized}, we also evaluate the effectiveness of \textit{GenDD} under data imbalance using the ImageNet-LT benchmark~\citep{liu2019large}.
We train ResNet models for 90 epochs using \textit{RandomResizedCrop} and horizontal flipping as standard preprocessing. Following previous work~\citep{cui2024decoupled}, we report top-1 accuracy across Many-shot, Medium-shot, Few-shot, and All classes to comprehensively assess performance.

As shown in Table~\ref{tab:imagenetlt} and Table~\ref{tab:imagenetlt_detail} (Appendix~\ref{sec:imagenetlt_detail}), we observe an interesting phenomenon: \textit{GenDD without label supervision can even achieve slightly better performance than IKL-KD incorporating labels. However, there are few accuracy gains after applying the label supervision with Distribution Contraction, which is a different behaviour compared to balanced data.} 
This phenomenon gives us new insight into KD for imbalanced data: the necessity of labels for KD on imbalanced data. As this work focuses on generative learning of KD, we leave this problem as our future work.

\begin{figure*}[tb!]
    \centering
    \subfloat[\textit{Split Tokenization} for token dimension]            { \includegraphics[width=0.5\linewidth]
     {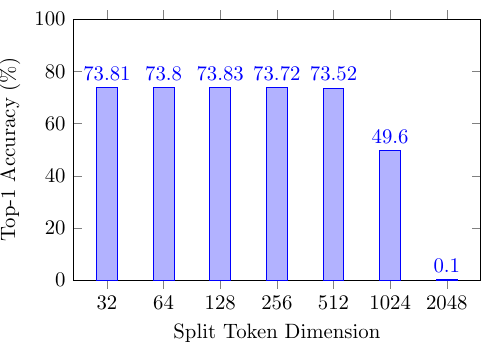}  \label{fig:split_tokenization_ablation}}
    \subfloat[$1-\lambda$ for \textit{Distribution Contraction}]            { \includegraphics[width=0.5\linewidth]
    {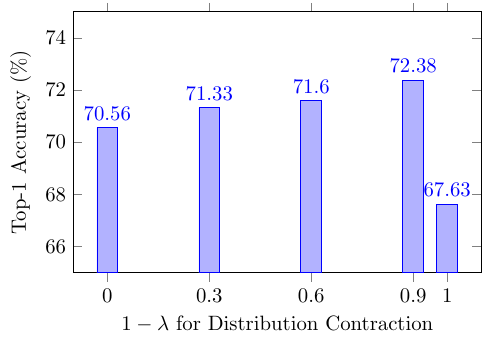} \label{fig:distribution_contraction_ablation}}
    \caption{
        \textbf{Ablation studies on \textit{Split Tokenization} and \textit{Distribution Contraction}.}
        (a) Top-1 Accuracy with different token dimension for \textit{Split Tokenization}. The teacher-student configuration of (ResNet-50, MobileNet) is used on ImageNet.;
        (b) Top-1 Accuracy under different $1-\lambda$ values for \textit{Distribution Contraction}. The teacher-student configuration of (ResNet-34, ResNet-18) is used on ImageNet.
    }
\end{figure*}

\subsection{Ablation Studies}
\label{sec:exp_ablation}
{\bf Ablation of \textit{Split Tokenization}.}
We validate the necessity of the proposed \textit{Split Tokenization} under the (ResNet-50, MobileNet) teacher-student configuration on ImageNet. As shown in Figure~\ref{fig:split_tokenization_ablation}, the model maintains competitive performance when the token dimension is $\leq 256$. However, accuracy drops sharply to $0.1\%$ as the token dimension increases from $512$ to $2048$, highlighting the high-dimensional optimization challenge and the effectiveness of \textit{Split Tokenization} in mitigating it.

{\bf Ablation on $\lambda$ for \textit{Distribution Contraction}.}
We validate the effectiveness of the proposed \textit{Distribution Contraction} technique under the (ResNet-34, ResNet-18) teacher configuration on ImageNet. As illustrated in Figure~\ref{fig:distribution_contraction_ablation}, \textit{GenDD} achieves competitive performance in the unsupervised KD setting with $1-\lambda=0.0$.
Deploying \textit{Distribution Contraction} technique with a proper $1-\lambda=0.9$, our model achieves significant performance gains. Interestingly, we observe that the model achieves much worse accuracy when the sample features contract to class centers, which indicates the importance of the continuity of the sample feature space. Inspired by this phenomenon, we apply the unsupervised mixup for diffusion training.

\begin{figure*}[tb!]
    \centering
    \subfloat[\textit{AdamW vs. SGD}]            { \includegraphics[width=0.5\linewidth]
     {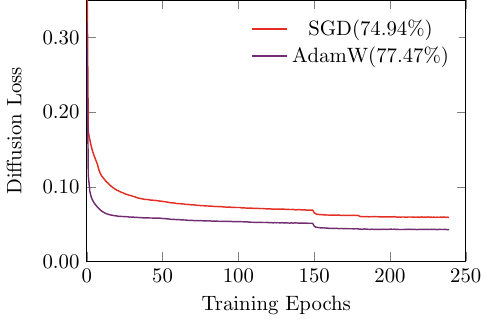}  \label{fig:sgd_adamw_ablation}}
    \subfloat[\textit{Step vs. Cosine LR}]            { \includegraphics[width=0.5\linewidth]
    {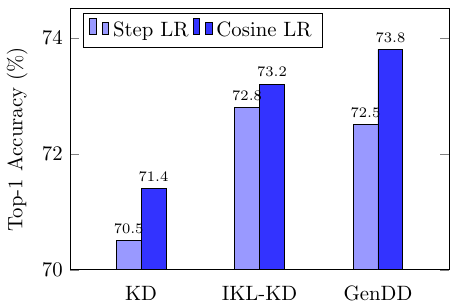} \label{fig:step_cos_lr_ablation}}
    \caption{
        \textbf{Ablation studies on optimizer and learning rate schedule.}
        (a) Comparison between AdamW and SGD optimizer for GenDD with teacher-student configuration of (ResNet-32x4, ResNet-8x4) on CIFAR;
        (b) Comparison between Step and Cosine learning rate schedule for GenDD with teacher-student configuration of (ResNet50, MobileNet) on ImageNet.
    }
\end{figure*}

{\bf AdamW vs. SGD Optimizer.}
We investigate the impact of different optimizers on training \textit{GenDD}. While SGD is commonly used for CNNs and AdamW/Adam are standard choices for Transformers, we adopt AdamW for \textit{GenDD} following previous work MAR~\citep{liu2023visual} and DDPM~\citep{ho2020denoising}. As shown in Figure~\ref{fig:sgd_adamw_ablation}, our empirical results demonstrate that AdamW leads to more stable and effective optimization for \textit{GenDD}.

{\bf Cosine vs. Step Learning Rate Schedule for GenDD.}
We study the impact of learning rate schedules on \textit{GenDD}, focusing on step decay and cosine annealing strategies. For fair comparison, we adopt the step schedule on CIFAR and reproduce the results of previous work with a cosine learning rate in their open-sourced code on ImageNet. Our empirical results show that the cosine schedule is critical for \textit{GenDD}, particularly on large-scale datasets such as ImageNet. As illustrated in Figure~\ref{fig:step_cos_lr_ablation}, cosine learning rates significantly accelerate convergence and improve overall performance.

\section{Conclusion, Limitation, and Future Work}
\label{sec:conclusion}
In this paper, we propose the \textit{Generative Distribution Distillation (GenDD)} algorithm, formulating the knowledge distillation (KD) as a conditional generation problem. Straightforward \textit{GenDD} pipeline suffers from the high-dimensional optimization disaster and the lack of label supervision. We propose the \textit{Split Tokenization} and \textit{Distribution Contraction} techniques correspondingly to address the above issues. With theoretical analysis, we prove that \textit{GenDD} with \textit{Distribution Contraction} approximates the multi-task learning (combining the reconstruction loss and the cross-entropy loss), while eliminating the multi-step sampling during training and achieving efficient optimization. Experimental results in unsupervised/supervised KD demonstrate the effectiveness of our method.

At inference, we adopt a 64-step sampling to generate image representations for classification, which can cause slightly higher latency. We would explore the few-step diffusion models or flow matching to overcome this limitation in future work.

\bibliography{iclr2025_conference}
\bibliographystyle{iclr2025_conference}

\newpage
\section{Appendix}
\label{sec:appendix}

\subsection{Proof of Theorem~\ref{thm:thm_gendd}}
\label{sec:proof_theorem1}

We begin by restating the \textit{GenDD} objective and the \textit{multi-task} objective. The \textit{GenDD} objective is the expected noise prediction loss over noisy data \( \tilde{x}_m \):
\begin{equation}
    \mathcal{L}_{\text{GenDD}} = \mathbb{E}_{x, m, \epsilon} \left[ \left\| \epsilon - \epsilon_{\theta}(\tilde{x}_m, m, \mathcal{F}_{s}(x)) \right\|^2 \right],
    \label{eq:gendd_loss}
\end{equation}
where \( \tilde{x}_m \) is given by:
\begin{equation}
    \tilde{x}_m = \sqrt{\bar{\alpha}_m} \tilde{x}_{\text{0}} + \sqrt{1 - \bar{\alpha}_m} \epsilon, \quad  \tilde{x}_{\text{0}} = \lambda \hat{x}_{0} + (1 - \lambda) c_y \ \textbf{(by Definition~\ref{thm:distribution_contraction})}, 
    \label{eq:gendd_forward}
\end{equation}
where \( \hat{x}_{0} \) is the teacher feature and \( c_y \) is the class center for class \( y \).

The multi-task objective consists of two parts:
- The reconstruction loss via noise prediction:
\[
\mathcal{L}_{\text{noise}} = \mathbb{E}_{x, m, \epsilon} \left[ \left\| \epsilon - \epsilon_{\theta}(\hat{x}_m, m, \mathcal{F}_s(x)) \right\|^2 \right],
\]
- The classification loss via cross-entropy:
\[
\mathcal{L}_{\text{CE}} = \mathbb{E}_{x} \left[ \mathcal{L}_{\mathrm{CE}} \left( \mathcal{C}_{t}(\hat{x}_0^{'}), y \right) \right],
\]
where \( \hat{x}_m \) is the noisy version of \( \hat{x}_{0} \), \( \mathcal{C}_{t} \) is the teacher classifier, and $\hat{x}_{0}^{'}$ is the generated feature.

Thus, the multi-task loss is:
\[
\mathcal{L}_{\text{multi}} = \gamma_{0} \mathcal{L}_{\text{noise}} + \gamma_{1} \mathcal{L}_{\text{CE}},
\]
where \( \gamma_{0} \) and \( \gamma_{1} \) are scaling constants.


{\bf Gradients of $\mathcal{L}_{GenDD}$ Regarding $\hat{x}_{0}^{'}$}. 

With the single-step estimation of $\hat{x}_{0}^{'}$,
\begin{equation}
    \hat{x}_{0}^{'} = \frac{1}{\sqrt{\bar \alpha_{m}}}(\tilde{x}_{m}-\sqrt{1-\bar \alpha_{m}} \epsilon_{\theta}).
    \label{eq:single_step_estimation}
\end{equation}
Then substitute $\tilde{x}_{m}$ in Eq.~\ref{eq:gendd_forward},
\begin{eqnarray}
    &\hat{x}_{0}^{'} = \frac{1}{\sqrt{\bar \alpha_{m}}}(\sqrt{\bar \alpha_{m}} \tilde{x}_{0} + \sqrt{1-\bar \alpha_{m}} \epsilon -\sqrt{1-\bar \alpha_{m}} \epsilon_{\theta}) = \tilde{x}_{0} + \frac{\sqrt{1-\bar \alpha_{m}}}{\sqrt{\bar \alpha_{m}}}(\epsilon - \epsilon_{\theta}), \\
    &\epsilon - \epsilon_{\theta} = \sqrt{\frac{\bar \alpha_{m}}{1-\bar \alpha_{m}}}(\hat{x}_{0}^{'} - \tilde{x}_{0})
\end{eqnarray}

With Eq.~\ref{eq:single_step_estimation},
\begin{eqnarray}
    \mathcal{L}_{GenDD} &=&  \mathbb{E}_{x, m, \epsilon} \left[ \left\| \epsilon - \epsilon_{\theta}(\tilde{x}_m, m, \mathcal{F}_{s}(x)) \right\|^2 \right] \\
    &=&  \mathbb{E}_{x, m, \epsilon} \left[ \left\| \epsilon - \frac{1}{\sqrt{1-\bar \alpha_{m}}}(\tilde{x}_{m} - \sqrt{\bar \alpha_{m}} \hat{x}_{0}^{'}) \right\|^2 \right] \\
    &=& \mathbb{E}_{x, m, \epsilon} \left[ \| \epsilon - \epsilon_{\theta}(\hat{x}_{0}^{'}) \|^{2} \right].
\end{eqnarray}
We take gradients of $\mathcal{L}_{GenDD}$ with respect to $\hat{x}_{0}^{'}$,
\begin{eqnarray}
    \nabla_{\hat{x}_{0}^{'}} \mathcal{L}_{GenDD} &=&  2\sqrt{ \frac{\bar \alpha_{m}}{1-\bar \alpha_{m}}} (\epsilon - \epsilon_{\theta}) \\
    &=& \frac{2\bar \alpha_{m}}{1-\bar \alpha_{m}} (\hat{x}_{0}^{'} - \tilde{x}_{0}) \\
    &=& \frac{2\bar \alpha_{m} }{1-\bar \alpha_{m}} (\hat{x}_{0}^{'} -\lambda \hat{x}_{0} -(1-\lambda) c_{y}) \\
    &=& \left[ \frac{2\bar \alpha_{m}}{1-\bar \alpha_{m}} (\hat{x}_{0}^{'} - \hat{x}_{0}) \right] + \left[  \frac{2 (1-\lambda) \bar \alpha_{m}}{1-\bar \alpha_{m}} (\hat{x}_{0} - c_{y}) \right].
    \label{eq:gradient_gendd}
\end{eqnarray}

{\bf Gradients of $\mathcal{L}_{multi}$ Regarding $\hat{x}_{0}^{'}$.}

Similar to $\mathcal{L}_{GenDD}$, the gradient of $\mathcal{L}_{noise}$ regrading $\hat{x}_{0}^{'}$ is:
\begin{equation}
    \nabla_{\hat{x}_{0}^{'}} \mathcal{L}_{noise} = \frac{2\bar \alpha_{m}}{1-\bar \alpha_{m}} (\hat{x}_{0}^{'} - \hat{x}_{0})
\end{equation}

For the $\mathcal{L}_{CE}$ item, we assume the \textit{weight} of the frozen linear classifier $\mathcal{C}_{t}(\cdot)$ is $W \in \mathbb{R}^{C \times d}$, where $d$ is the feature dimension and $C$ is the number of interested classes. Then,
\begin{equation}
    \mathcal{L}_{CE} = -\log p(y|\hat{x}_{0}^{'}), \ \text{where} \  p(y|\hat{x}_{0}^{'}) = \frac{W_{y} \hat{x}_{0}^{'}}{\sum_{i=1}^{C} W_{i} \hat{x}_{0}^{'}},
\end{equation}
\begin{eqnarray}
    \nabla_{\hat{x}_{0}^{'}} \mathcal{L}_{CE} = \sum_{i=1}^{C} \left[p(i|\hat{x}_{0}^{'}) W_{i} \right] - W_{y}.
\end{eqnarray}
Overall, the gradients of $\mathcal{L}_{multi}$ regarding $\hat{x}_{0}$ is derived as:
\begin{equation}
    \nabla_{\hat{x}_{0}^{'}} \mathcal{L}_{multi} = \gamma_{0} \left[ \frac{2 \bar \alpha_{m}}{1-\bar \alpha_{m}} (\hat{x}_{0}^{'} - \hat{x}_{0}) \right] + \gamma_{1} \left[ (\hat{x}_{0}^{''} - c_{y}) \right], 
    \label{eq:gradient_multitask}
\end{equation}
where $\hat{x}_{0}^{''} = \sum_{i=1}^{C}\left[ p(i|\hat{x}_{0}^{'}) W_{i}\right]$, and $c_{y}=W_{y}$. Note that we also use $c_{y} = W_{y}$ in \textbf{Definition}~\ref{thm:distribution_contraction}.

In Eq.~\ref{eq:gradient_multitask}, we observe that $\hat{x}_{0}^{''} \approx \hat{x}_{0}^{'}$ when the predicted probability $p(y|\hat{x}_{0}^{'})$ approaches 1.0. To ensure consistency between training and inference in multi-task learning, it is crucial to employ multi-step sampling to obtain accurate estimates of $\hat{x}_{0}^{'}$ during the optimization of $\mathcal{L}_{\text{CE}}$. In such cases, the improved quality of $\hat{x}_{0}^{'}$ naturally leads to higher classification confidence, making the approximation $\hat{x}_{0}^{''} \approx \hat{x}_{0}^{'}$ practically valid.

{\bf Summary.} Therefore, by comparing Eq.\ref{eq:gradient_gendd} and Eq.\ref{eq:gradient_multitask}, we conclude that $\nabla_{\hat{x}{0}^{'}} \mathcal{L}_{\text{GenDD}} \approx \nabla_{\hat{x}_{0}^{'}} \mathcal{L}_{\text{multi}}$. This indicates that GenDD, augmented with the distribution contraction mechanism, functions as a gradient-level surrogate for multi-task learning, without requiring the explicit application of the classifier loss to $\hat{x}_{0}^{'}$.

\subsection{Comparisons of Sensitivity to Hyperparameters with Previous Methods on ImageNet}
\label{sec:comparison_hyper_parameter}
\begin{table}[h]
    \centering
    \caption{\textbf{Sensitivity to hyperparameters.} Advanced KD methods often involve complex hyperparameter tunning. Our \textit{GenDD} method consistently works well across diverse teacher-student configurations with $\lambda=0.9$ on ImageNet. }
    \begin{tabular}{ccc}
         \toprule
         Method& Teacher-student  &Hyperparameter configuration \\
         \midrule
         \multirow{2}{*}{KD} &ResNet-34 --- ResNet-18 & \multirow{2}{*}{$w_{kl}=0.5$, $w_{ce}=0.5$, $T=1.0$}\\
                            &ResNet-50 --- MobileNet & \\
         \midrule
         \multirow{2}{*}{DKD} &ResNet-34 --- ResNet-18 &$w_{ce}=1.0$, $w_{\alpha}=1.0$, $w_{\beta}=0.5$, $T=1.0$ \\
                             &ResNet-50 --- MobileNet &$w_{ce}=1.0$, $w_{\alpha}=1.0$, $w_{\beta}=2.0$, $T=1.0$ \\
         \midrule
         \multirow{2}{*}{IKL-KD} &ResNet-34 --- ResNet-18 &$w_{ce}=1.0$, $w_{\alpha}=1.0$, $w_{\beta}=0.5$, $T=1.0$  \\
                                &ResNet-50 --- MobileNet &$w_{ce}=1.0$, $w_{\alpha}=4.0$, $w_{\beta}=1.0$, $T=1.0$ \\
         \midrule
         \multirow{2}{*}{GenDD}  &ResNet-34 --- ResNet-18 &\multirow{2}{*}{$\lambda$=0.9} \\
                                &ResNet-50 --- MobileNet & \\
        \bottomrule
    \end{tabular}
    \label{tab:my_label}
\end{table}

\subsection{Detailed Performance on Few-, Medium-, and Many-shot}
\label{sec:imagenetlt_detail}
\begin{table}[h]
    \centering
    \caption{\textbf{Top-1 accuracy(\%) on the ImageNet-LT validation with GenDD.} ``*'' represents the unsupervised setting.}
    {
    \begin{tabular}{lcccccc}
      \toprule
      Method & Teacher &Student &All &Few &Medium &Many \\
      \midrule
      Baseline &- &ResNet-18 &63.16 &33.47 &5.88 &41.15\\
      Baseline &- &ResNet-50 &67.25 &38.56 &8.21 &45.47 \\
      \midrule 
      KD &ResNeXt-101 &ResNet-18 &64.60 &37.88 &9.53 &44.32 \\
      KD &ResNeXt-101 &ResNet-50 &68.83 &42.31 &11.37 &48.31 \\
      IKL-KD &ResNeXt-101 &ResNet-18 &66.60 &38.53 &8.19 &45.21 \\
      IKL-KD &ResNeXt-101 &ResNet-50 &70.06 &43.47 &10.99 &49.29 \\
      \midrule
      \textbf{GenDD*} &ResNext-101 &ResNet-18 &66.71 &39.02 &8.66 &45.54\\
      \textbf{GenDD*} &ResNeXt-101 &ResNet-50 &70.12 &43.52 &10.84 &49.31\\
      \bottomrule
    \end{tabular}
    }
    \label{tab:imagenetlt_detail}
  \end{table}

\end{document}